# StemVLA:An Open-Source Vision-Language-Action with Future 3D Spatial Geometry Knowledge and 4D Historical Representation

Jiasong Xiao
*Image Processing*
*Ricoh Software Research Center (Beijing) Co., Ltd.*
Beijing, China
Jiasong.Xiao@cn.ricoh.com

Yutao She
*Guanghua School of Management*
*Peking University*
Beijing, China
ytshe25@stu.pku.edu.cn

Kai Li
*Image Processing*
*Beijing YZH Engineering Technology Co., Ltd*
Beijing, China
likai@bj-yzh.com

Yuyang Sha
*Faculty of Applied Sciences*
*Macao Polytechnic University*
Macao, China
p2215539@mpu.edu.mo

Zi'ang Cheng
*School of Artificial Intelligence*
*Tianjin University*
Tianjin, China
3023244167@tju.edu.cn

***Abstract*—Vision-language-action (VLA) models integrate visual observations and language instructions to predict robot actions, demonstrating promising generalization in manipulation tasks. However, most existing approaches primarily rely on direct mappings from 2D visual inputs to action sequences, without explicitly modeling the underlying 3D spatial structure or temporal world dynamics. Such representations may limit spatial reasoning and long-horizon decision-making in dynamic environments. To address this limitation, we propose StemVLA, a novel framework that explicitly incorporates both future-oriented 3D spatial knowledge and historical 4D spatiotemporal representations into action prediction. First, instead of relying solely on observed images, StemVLA forecasts structured 3D future spatial-geometric world knowledge, enabling the model to anticipate upcoming scene geometry and object configurations. Second, to capture temporal consistency and motion dynamics, we feed historical image frames into a pretrained video-geometry transformer backbone to extract implicit 3D world representations, and further aggregate them across time using a temporal attention module, termed VideoFormer [20], forming a unified 4D historical spatiotemporal representation. By jointly modeling 2D observations, predicted 3D future structure, and aggregated 4D temporal dynamics, StemVLA enables more comprehensive world understanding for robot manipulation. Extensive experiments in simulation demonstrate that Stem-VLA achieves an average accuracy of 92.0% across the LIBERO subsets, and 86.0% on the long-horizon LIBERO-Long subset.***

## I. INTRODUCTION

Embodied AI has recently demonstrated substantial progress in language-conditioned robotic manipulation [1-4, 8, 9, 11], enabling agents to execute diverse tasks in complex and dynamic environments [5, 6, 7, 10]. Among various paradigms, Vision-Language-Action (VLA) technologies have emerged as a prominent direction by integrating visual perception, language understanding, and low-level control within a unified framework. Leveraging the reasoning capabilities of Multimodal Large Language Models (MLLMs) [12, 13], VLA systems allow robots to interpret high-level instructions and translate them into executable action sequences grounded in visual observations.

Despite these advances, most existing VLA approaches [19, 21-25] primarily rely on implicit 2D visual representations when mapping observations to action sequences. While such designs have achieved encouraging results, they often lack explicit modeling of structured 3D spatial geometry and long-range temporal dynamics, which are crucial for robust decision-making in dynamic environments. To enhance world understanding, several recent works [1, 2, 5, 7, 8, 11] attempt to jointly model image prediction and action prediction within a unified architecture, encouraging interaction between visual forecasting and control reasoning. Although promising, these approaches present several challenges. First, predicting full-resolution future frames introduces substantial redundancy, as large portions of predicted pixels overlap with current observations. Second, future spatial geometry is typically modeled only at the pixel level, without explicitly capturing structured 3D representations such as depth or scene layout [1, 9, 17, 18, 26, 42, 47, 49]. Third, historical observations are frequently encoded in a frame-wise manner, limiting the modeling of coherent spatiotemporal dynamics across multiple steps [2, 14, 25, 27, 46]. Finally, heavy reliance on high-level semantic embeddings may obscure fine-grained physical details necessary for precise manipulation.

To address these limitations, we propose StemVLA, a framework that explicitly incorporates structured world knowledge into the VLA architecture. StemVLA introduces two key components: (1) a 3D future spatial-geometric world knowledge prediction module embedded within the language model to anticipate upcoming scene geometry, and (2) a 4D historical spatiotemporal aggregation mechanism that fuses latent 3D features extracted from past observations. Together, these components enable the model to reason over both anticipated future structure and accumulated temporal context. Notably, both modules operate on implicit 3D representations extracted from visual inputs, avoiding the need for redundant explicit spatial supervision while preserving rich geometric and physical details.

Concretely, we employ the VGGT 3D reconstruction model [36] to extract latent 3D features from both historical observations and predicted future frames. These intermediate representations encode structured geometric information, such as depth and spatial layout, providing an implicit yet expressive characterization of scene structure. The extracted 3D features serve as supervisory signals for future world knowledge prediction, enabling the model to anticipate upcoming geometric configurations rather than merely extrapolating pixel values. To effectively integrate spatial information across time, we introduce a temporal aggregation module termed VideoFormer [20], which facilitates interact-ion among multi-dimensional spatial features along the temporal dimension and yields a unified 4D spatiotemporal representation for action prediction. By operating on latent 3D features instead of raw pixels, the proposed framework preserves both high-level semantic cues and fine-grained physical details, thereby improving the modeling of dynamic interactions in long-horizon manipulation tasks. Empirically, StemVLA demonstrates consistent and significant improvements across challenging robotic manipulation benchmarks. In terms of task success, StemVLA improves the overall success rate from 78.1% to 92.0%, with particularly notable gains in long-horizon and spatially complex scenarios. These results validate that explicitly incorporating future-oriented 3D geometric knowledge and historical 4D spatiotemporal representations substantially enhances action prediction and decision-making robustness.

## II. RELATED WORKS

### *A. Vision-Language-Action Models*

The field of Vision-Language-Action (VLA) was initially established by works that merged pre-trained visual and linguistic representations with policies conditioned on specific tasks for robotic control [2, 10, 30, 31]. Driven by the technological advancements in Large Language Models [32, 33] and Multimodal Large Language Models (MLLMs) [29, 34], this research direction has gained significant momentum through the creation of large-scale robot-centric datasets [7, 35, 37]. Among these, notable early achievements in the VLA field include the RT series models [2, 7, 38] and OpenVLA [1]. Specifically, related studies [1, 8, 15, 18, 30, 39, 41] have demonstrated that fine-tuning Multimodal Large Language Models (MLLMs) on robotic trajectory data enables these models to achieve excellent accuracy and generalization performance. Subsequent research has further proposed a variety of sophisticated methods, continuing to drive performance improvements in this field.

At the same time, another research stream focuses on the technical application of diffusion models and conducts in-depth investigations by leveraging their unique advantages in multimodal distribution modeling tasks. A number of studies [42-45] have designed task-specific network architectures that enable iterative denoising of action sequences based on sensory inputs, language commands, and embodied prior knowledge. A well-recognized limitation of such end-to-end paradigms that directly output actions is the lack of explicit chain-of-thought reasoning capability comparable to that of LLMs and VLMs. To address this core issue, many existing technical approaches [20-23] introduce an independent image or video generation model to achieve accurate prediction of future states, and then formulate the specific execution path of actions guided by these generated visualized goals.

However, this two-stage process often incurs considerable latency. To address this issue, several studies [1, 2, 9, 14, 17, 18, 25] have proposed integrated frameworks that consolidate the two tasks of future frame forecasting and action decision-making into a single model, aiming to harness the synergistic relationship between prediction and planning. Despite these advancements, the current state-of-the-art methods [26, 27, 42, 46, 47, 48, 49] in the field of Vision-Language-Action (VLA) are still confronted with a pressing challenge: the lack of 3D spatial geometry information and 4D spatiotemporal information.

### *B. Vision Representations for Robotics*

The Visual representation acts as the "eyes" of Vision-Language-Action (VLA) models, converting raw pixel data into semantically rich abstract features critical for grounding linguistic instructions in the physical world and enabling accurate action decisions. Early studies on embodied VLA models [1, 2, 9, 14, 41] primarily relied on 2D visual representations from planar images. Generated by pre-trained vision encoders, these representations encode texture, color, and luminance to support environmental perception and action planning. Nevertheless, 2D visual representations have inherent limitations for robotic applications: (1) lacking geometric, depth, and layout information, hindering perception of 3D spatial relationships essential for robotic manipulation; (2) high sensitivity to appearance variations from viewpoint changes, lighting, and occlusion, which may destabilize performance; (3) inability to support high-precision physical interaction tasks requiring accurate spatial reasoning.

Visual representations have evolved from 2D images to 3D spatial structures, and recently to 4D spatiotemporal dynamic features. Our StemVLA model integrates 3D spatial-geometric representations into VLA frameworks, enhancing generalization and manipulation performance via geometric and positional awareness from 2D inputs. Despite these advantages, 3D spatial-geometric representations function as static "snapshots" of the external world, lacking dynamic information critical for dynamic perception, motion forecasting, event causal reasoning, and long-horizon planning. Thus, StemVLA learns 4D historical spatiotemporal representations from video sequences, integrating 3D spatial and temporal information to offer key benefits: (1) perceiving motion states and variations; (2) inherent temporal context endows predictive capabilities for future motion and scene dynamics, essential for planning; (3) analyzing spatiotemporal sequences improves causal reasoning via chronological event reasoning.

## III. METHODOLOGY

### *A. Problem Definition and Notation*

This study aims to advance the evolution of representational forms from 2D images to 4D spatiotemporal features, thereby boosting the model's spatiotemporal comprehension and action prediction capabilities for complex, continuous robotic manipulation tasks and ultimately enabling dual enhancements in its end-to-end performance and generalization capability. We propose a robot manipulation-oriented Vision-Language-Action (VLA) framework that fuses 2D images, 3D spatial-geometric information, 4D spatiotem-poral information, and linguistic instructions. Using the reasoning and generalization capabilities of large language models (LLMs), our method generates dense, sequential action trajectories in a 4D action space.

To aggregate 4D historical spatiotemporal representations from 2D image data, we first employ the VGGT Aggregator [36] to encode 2D historical images and extract latent 3D spatial-geometric features $f_\tau^{3D}$. These 3D representations gen-erated by the VGGT Aggregator [36] are inherently high-dimensional, as they integrate image feature sequences across diverse viewpoints. To enable efficient cross-dimensional aggregation of representations across spatial, temporal, and multi-view dimensions, we introduce the History Aggregator (HA), i.e., VideoFormer [20], which fuses the aforementioned information into 4D historical spatiotemporal representations $f_\tau^{4D}$.

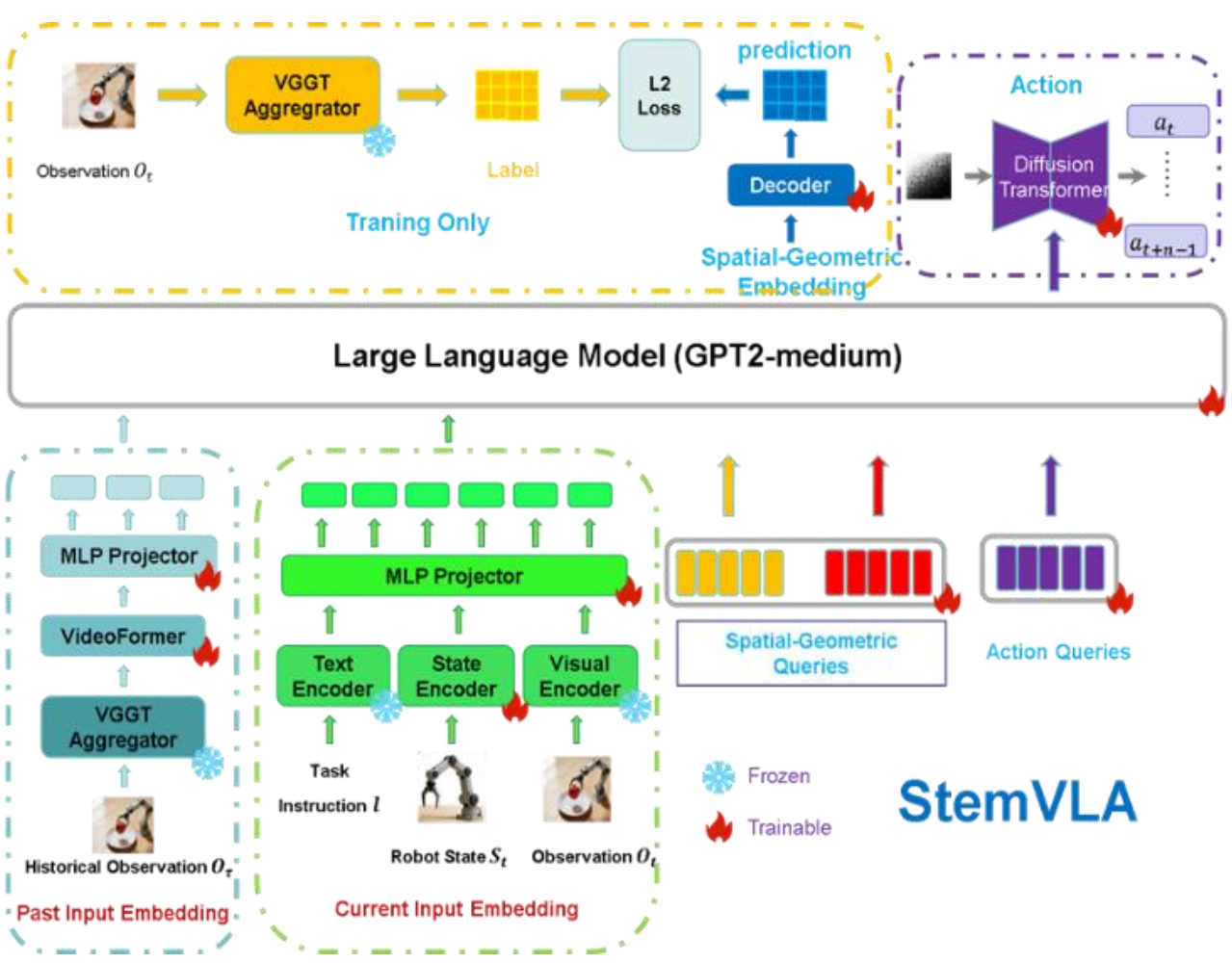

**Figure 1: Framework Overview**. Given the historical observation $o_\tau$, the current language Instruction $l$, robot state $s_t$, and Observation $O_t$, StemVLA employs the VGGT Aggregator [36] and VideoFormer [20] to encode historical observations, and uses frozen text and visual encoders along with a trainable state encoder to encode the current multimodal input. These tokens, together with a learnable set of <spatial-geometric> queries, are processed by a large language model to produce spatial-geometric world knowledge. A separate <action> query generates a latent action embedding, which serves as the conditional input to a diffusion transformer, refining gaussian noise into an n-step action sequence $\hat{a}_{t:t+n-1}$, The prediction heads highlighted by the dashed box are used only during the training phase; during inference, these heads are skipped and operations are performed directly on the world embedding.

$$f_\tau^{3D} = \text{VGGT_Aggregator}(i_\tau) \quad (1)$$

$$f_\tau^{4D} = \text{VideoFormer}\left(f_\tau^{3D}\right) \quad (2)$$

Where $\tau \in (t-T, t]$, $T$ denotes historical time window length and $t$ denotes the current time step.

At each time step $t$, our model processes a set of heterogeneous inputs: a natural language instruction $l$, a raw visual image observation $i_t$, a 4D historical spatiotemporal feature $f_\tau^{4D}$ and the robot's proprioceptive state $s_t$. To integrate 3D spatial-Geometric information, we introduce a learnable <spatial-geometric> query. All utilized inputs are spatially aligned to construct a unified token sequence. The Multimodal Large Language Model (MLLM) then fuses this sequence into a compact latent representation, termed the spatial embedding.

$$w_{t+n} = \text{MLLM}(f_\tau^{4D}, l, s_t, \hat{i}_t, \text{spatial-geometric}) \quad (3)$$

Subsequently, the spatial embedding is fed into the 3D Future Spatial-Geometric World Knowledge Predictor (FSGWP) to predict the 3D Future Spatial-Geometric World Knowledge under 3D Future Spatial-Geometric Supervision. Specifically, the FSGWP extrapolates the world state n steps into the future.

$$\hat{p}_{t+n} = \text{FSGWP}(w_{t+n}) \quad (4)$$

Conditioned on the spatial embedding $w_{t+n}$, the unified model M processes the <action> query to construct a latent action embedding that aggregates task-relevant information. This compressed representation is then passed to a Denoising-Diffusion Transformer (DiT) [42], which generates the complete n-step action sequence via iterative refinement.

$$\hat{a}_{t:t+n-1} = \text{DiT}(\text{MLLM}(f_\tau^{4D}, l, s_t, \hat{i}_t, \text{action})) \quad (5)$$

## B. Model Architecture

As shown in Figure 1, we present StemVLA, a unified Transformer-based architecture designed for robotic manipulation. The model consists of four core components: Modality-Specific Inputs, a Shared MLLM Backbone, Training-phase-specific 3D Future Spatial-Geometric Supervision, and Action Outputs.

**Modality-Specific Inputs.** This model processes heterogeneous inputs including natural language instructions $l$, 2D image observations $i_t$, proprioceptive states $s_t$ and Video $v$ with dedicated encoders. We encode linguistic inputs using the CLIP text encoder, encode 2D image streams using a Masked AutoEncoder to extract fine-grained representations such as color and texture and proprioceptive signals are processed via a combination of an attention module and fully-connected layers. For video sequences, we construct a hybrid architecture that integrates the VGGT Aggregator [36] and VideoFormer [20], termed the Historical Spatio Temporal Encoder which fuses the aforementioned information to generate 4D historical spatiotemporal representations. Specifically, the VGGT Aggregator [36] is responsible for extracting visual spatial-geometric features, while Video-Former [20] is tasked with the task of fusing these spatial-geometric features across the temporal dimension. This synergistic design enables this encoder to capture both fine-grained spatial structures and long-range temporal dependencies. Finally, we embed two learnable query vectors, <spatial-geometric> and <action>, into the encoded representations. The <spatial-geometric> query vector enables the VLA to predict 3D future spatial-geometric information, while the <action> query vector aggregates task-relevant information and predicts future action sequences and control commands. Together, they form a complementary and synergistic dual-query mechanism that supports the VLA's integrated spatiotemporal and action reasoning capabilities.

**Shared MLLM Backbone.** A GPT-2-based [51] variant of type of Multimodal Large Language Models (MLLMs) acts as the fusion backbone for integrating information across modalities and query tokens. This backbone enables effective aggregation of low-level 2D visual features (such as color, texture), middle-level 3D spatial-geometric features (such as geometric and positional knowledge), and high-level coherent 4D historical spatiotemporal representations (such as event sequences). These multi-level representations jointly capture comprehensive world knowledge.

**Training-phase-specific 3D Future Spatial-geometric Supervision**. Its core function is to constrain the model to learn precise spatial-geometric representations through explicit spatial supervision. It comprises three sub-modules: 1) The Label Generation Module receives the visual observation data $O_t$ at future timestamps, feeds it into the VGGT Aggregator [36] for spatial-geometric feature extraction and modeling, and generates fine-grained ground-truth labels for the 3D future spatial-geometric environment.

These labels capture the target geometric structure of future scenes that the model is to learn. 2) The Synchronous Output Module receives multimodal fusion representations from the backbone network, then incorporates world knowledge embeddings, decodes the fused representations into predicted 3D future spatial-geometric environment information, and enables explicit prediction of future scene geometric structure. 3) The 3D Future Spatial-Geometry Supervision Module computes the error between the predicted 3D spatial-geometric information and the ground-truth labels generated by the VGGT Aggregator [36] using the L2 loss function. This module establishes a 3D future spatial supervision mechanism, which constrains the model to learn more precise spatial-geometric features, and thereby enhances the spatial rationality of downstream action generation tasks.

**Table 1. Experimental setup details for StemVLA**. This table summarizes the key configurations, including hardware specifications, optimizer choices, core hyperparameters, loss functions, datasets, and training cycles used in our main experiments and ablation studies.

| *Category* | *Configuration and Parameters* |
|---|---|
| Implementation & Hardware | Framework: PyTorch<br>Training Hardware:8 × NVIDIA A100 GPUs |
| Optimizer & Scheduler | Optimizer: AdamW<br>VLM Initial Learning Rate: 6e-4<br>Diffusion Initial Learning Rate: 1.5e-4<br>Weight Decay: 1e-4<br>Learning Rate Schedule: Cosine with 5% warm-up |
| Core Hyperparameters | Batch Size: 20<br>Diffusion Steps (in DiT): 10<br>Query Length (per modality): 9 |
| Loss Weights | Pixel-wise Loss: 0.1<br>3D Future Spatial-Geometry World Loss: 0.1<br>Action Loss: 1.0 |
| Ablation Study | Pre-training Dataset: LIBERO-90 Dataset<br>Fine-tuning Dataset and Evaluation Dataset: LIBERO-Long, LIBERO-Goal, LIBERO-Spatial and LIBERO-Object. |
| Training Cycle & Evaluation | Total Epochs: 40<br>Checkpoint Selection: The checkpoint with the highest validation success rate (SR) is selected for final evaluation. |

**Action Outputs.** A Denoising-Diffusion Transformer (DiT) [42] processes the <Action> embeddings to generate robot action sequences through iterative denoising. Collectively, this architecture unifies Modality-Specific Inputs and Synchronous Outputs within a 4D spatiotemporal framework, enabling the model to capture comprehensive world information and thereby achieve substantial improvements in performance and generalization.

### *C. 4D Historical Spatiotemporal Representation*

The 4D Historical Spatiotemporal Representation fuses spatial and temporal information to enable dynamic perception, motion forecasting, event sequences causal reasoning, and long-horizon action planning. By leveraging the VGGT Aggregator [36] to extract 3D spatial priors from 2D images and employing the History Aggregator (HA) to aggregate 3D spatial-geometric information over time, we construct a comprehensive 4D historical spatiotemporal representation.

### *D. 3D Future Spatial World Knowledge*

To enable StemVLA to truly and holistically understand 3D Future Spatial-Geometric World Knowledge, we employ the FSGWP, which is guided by the <Spatial> query to generate such world knowledge. This process resembles human learning: a holistic understanding of knowledge is achieved not through passive input alone, but through active reconstruction and externalization of input information. The loss associated with 3D Future Spatial-Geometric World Knowledge is formulated as

$$\mathcal{L}_{\text{FSGWP}} = \frac{1}{HW} \sum_{i,j} \left( \hat{p}_{t+n}^{(i,j)} - p_{t+n}^{(i,j)} \right)^2 \tag{6}$$

### *E. Action Generation via Diffusion*

StemVLA predicts action trajectories $\hat{a}_{t:t+n-1}$ through a diffusion process. The action prediction loss is formulated as:

$$\mathcal{L}_{\text{action}} = \mathbb{E}_{k,\varepsilon} \left\| \varepsilon - \varepsilon_\theta \left( \sqrt{\bar{\alpha}_k}\, a_{t:t+n-1} + \sqrt{1-\bar{\alpha}_k}\, \varepsilon, k, \mathbf{c} \right) \right\|_2^2 \tag{7}$$

## IV. EXPERIMENTS

To validate the effectiveness of the proposed StemVLA model in embodied AI tasks, we designed and implemented a comprehensive training and evaluation pipeline. As shown in Table 1, this process aligns with modern deep learning paradigms and incorporates advanced techniques, including the AdamW [50] optimizer and a cosine learning rate schedule. We conducted systematic main and ablation experiments on large-scale, multi-task datasets to thoroughly evaluate the model's generalization capability and the efficacy of its core components.

### *A. Main Experiments*

**Simulation Setup**

We perform our primary evaluations on the LIBERO [49] benchmark. LIBERO [49] is a simulation benchmark platform for research on robotic lifelong learning and cross-task knowledge transfer. It features four specialized test suites: LIBERO-Spatial, LIBERO-Object, LIBERO-Goal, and LIBERO-Long, each consisting of 10 custom-designed tasks. Each task includes 50 human teleoperated demonstration instances, which form the platform's foundational dataset. This platform enables comprehensive evaluation of a robot's multi-dimensional capabilities, including spatial relational reasoning, object-centric precise manipulation, goal-conditioned task accomplishment, and long-horizon continuous manipulation, serving as a critical benchmark for evaluating a model's knowledge reuse and cross-task transfer performance.

**Results**

As shown in Table 2, on the LIBERO benchmark [49], StemVLA delivers superior or comparable performance across all test tracks relative to prior approaches, which is attributed to its 3D future spatial-geometric world knowledge prediction mechanism.

### *B. Ablation Study*

To systematically evaluate the proposed method, this section is structured around addressing the following research questions. For the ablation study, we first pre-train

StemVLA on LIBERO-90 dataset, and subsequently fine-tune and evaluate the model on LIBERO-Long, LIBERO-Object, LIBERO-Spatial and LIBERO-Goal datasets.

Q1: The effectiveness of 4D Historical Spatiotemporal Representation for robot manipulation?

As shown in Table 3, the experimental results demonstrate that integrating the 4D Historical Spatiotemporal Representation yields a significant improvement in robotic manipulation performance. Models incorporating this representation consistently outperform their ablated counterparts across all four LIBERO subsets: on the LIBERO-Long dataset, accuracy improves from 83.5% to 86.0%; on the LIBERO-Object datasets, accuracy improves from 92.0% to 96.0%; on the LIBERO-Spatial datasets accuracy improves from 91.5% to 96.0%; and on the LIBERO-Goal datasets, accuracy improves from 90.5% to 92.0%. The 4D Historical Spatiotemporal Representation encapsulates both spatial and temporal information, and exhibits notable advantages in dynamic perception, motion forecasting, causal event sequence reasoning, and long-horizon action planning.

**Table 2**: **The extended LIBERO experiments**. StemVLA achieves the more competitive performance across all tracks compared to previous approaches. The best results are bolded

| **Methods** | **Scores(%)** | | | | **Ave** |
|---|---|---|---|---|---|
| | ***Spatial*** | ***Object*** | ***Goal*** | ***Long*** | |
| Diffusion Policy [42] | 78.3 | 92.5 | 68.3 | 50.5 | 72.4 |
| Octo [8] | 78.9 | 85.7 | 84.6 | 51.1 | 75.1 |
| OpenVLA [1] | 84.7 | 88.4 | 79.2 | 53.7 | 76.5 |
| SpatialVLA [17] | 88.2 | 89.9 | 78.6 | 55.5 | 78.1 |
| CoT-VLA [27] | 81.1 | 87.5 | 91.6 | 87.6 | 69.0 |
| StemVLA | 96.0 | 96.0 | 92.0 | 86.0 | 92.0 |

Q2: The effectiveness of 3D Future Spatial Geometry World Knowledge for robot manipulation?

To examine the impact of integrating 3D future spatial-geometric world knowledge into robot manipulation tasks, we removed the FSGWP module while holding all other experimental conditions constant. As shown in Table 3, the experimental results demonstrate that the FSGWP module's integration of camera intrinsic/extrinsic parameters, depth information, point cloud data, and trajectory tracking yields a significant improvement in robotic manipulation performance. Models incorporating the FSGWP module consistently outperform their ablated counterparts across all four subsets of the LIBERO dataset: on the LIBERO-Long dataset, accuracy improves from 67.0% to 86.0%; on the LIBERO-Object datasets, accuracy improves from 78.0% to 96.0%; on the LIBERO-Spatial datasets accuracy improves from 76.5% to 96.0%; and on the LIBERO-Goal datasets, accuracy improves from 72.0% to 92.0%. This unified representation enables simultaneous perception of 3D spatial structure, object motion trajectories, and spatial relationships, delivering richer geometric and dynamic information to support action planning. By fusing depth-based spatial constraints, point cloud geometry, and trajectory tracking-derived motion patterns, the model achieves enhanced precision in environmental understanding and more stable execution performance in complex manipulation tasks, thereby substantially boosting success rates for long-horizon manipulation operations.

**Table 3**: The Ablation experiments on LIBERO Datasets. Performance comparison among StemVLA without 3D Future Spatial Geometry World Knowledge, StemVLA without 4D Historical Spatiotemporal Representation, and intact StemVLA.

| **Method** | ***LIBERO-Long*** | ***LIBERO-Object*** | ***LIBERO-Spatial*** | ***LIBERO-Goal*** |
|---|---|---|---|---|
| StemVLA without 3D Future Spatial Geometry World Knowledge | 67.0% | 78.0% | 76.5% | 72.0% |
| StemVLA without 4D Historical Spatiotemporal Representation | 83.5% | 92.0% | 91.5% | 90.5% |
| StemVLA(intact) | 86.0% | 96.0% | 96.0% | 92.0% |

## V. LIMITATION & FUTURE WORKS

While StemVLA attains state-of-the-art results on the LIBERO [49] benchmark, several key limitations persist. The current model is restricted to parallel gripper manipulation and was trained in environments with limited geometric and material variability. Furthermore, the underlying DiT architecture occasionally generates jerky and sluggish motions, which undermines the fluency of real-time control. To miti-gate these limitations, our future research will focus on three core directions: (i) incorporating demonstrations for dexterous hand manipulation with detailed contact annotations; (ii) scaling up data collection and integrating on-policy fine-tuning to improve generalization and robustness on long-horizon tasks; and (iii) replacing the DiT architecture with more efficient Flow Matching techniques to substantially enhance motion smoothness and real-time control performance.

## VI. CONCLUSION

We present StemVLA, a novel Visual-Language-Action framework that integrates comprehensive representations, which is beneficial for temporal-related tasks. We address the critical limitations of lacking 3D spatial and 4D Historical Spatiotemporal Representations in embodied AI, which are vital for physical reasoning and future forecasting. What's more, StemVLA first introduces VGGT [36] as a 3D foundation model to extract an intermediate 3D spatial representation from the 2D images. Building upon this, the History Aggregator module further processes this representation to model temporal dynamics, resulting in a powerful 4D Historical Spatiotemporal Representation. Extensive experiments demonstrate the effectiveness of StemVLA, achieving a 92.0% average success rate on the LIBERO benchmark and outperforming many prior methods.